\documentclass[letterpaper]{article} % DO NOT CHANGE THIS
\usepackage{aaai25}  % DO NOT CHANGE THIS
\usepackage{times}  % DO NOT CHANGE THIS
\usepackage{helvet}  % DO NOT CHANGE THIS
\usepackage{courier}  % DO NOT CHANGE THIS
\usepackage[hyphens]{url}  % DO NOT CHANGE THIS
\usepackage{graphicx} % DO NOT CHANGE THIS
\usepackage{natbib}  % DO NOT CHANGE THIS AND DO NOT ADD ANY OPTIONS TO IT
\usepackage{caption} % DO NOT CHANGE THIS AND DO NOT ADD ANY OPTIONS TO IT
\usepackage{algorithm}
\usepackage{algorithmic}
\usepackage{adjustbox}
\usepackage{multirow}
\usepackage{booktabs}
\usepackage{color}
\usepackage{amsmath}
\definecolor{grey}{cmyk}{0,0,0,0.5} 
\newcommand*\rot{\rotatebox{90}}

\usepackage{newfloat}
\usepackage{listings}
\DeclareCaptionStyle{ruled}{labelfont=normalfont,labelsep=colon,strut=off} % DO NOT CHANGE THIS
\floatstyle{ruled}
\newfloat{listing}{tb}{lst}{}
\floatname{listing}{Listing}
\title{CP-DETR: Concept Prompt Guide DETR Toward Stronger \\Universal Object Detection}
\author{
    Qibo Chen\textsuperscript{\rm 1}\thanks{Corresponding Author}, 
    Weizhong Jin\textsuperscript{\rm 1}, 
    Jianyue Ge\textsuperscript{\rm 1}, 
    Mengdi Liu\textsuperscript{\rm 1}, 
    Yuchao Yan\textsuperscript{\rm 1},
    Jian Jiang\textsuperscript{\rm 1},
    Li Yu\textsuperscript{\rm 1},
    Xuanjiang Guo\textsuperscript{\rm 1}, 
    Shuchang Li\textsuperscript{\rm 1}, 
    Jianzhong Chen\textsuperscript{\rm 1}
}
\affiliations{
    \textsuperscript{\rm 1}China Mobile(Zhejiang) Research $\&$ Innovation Institute\\ 
    chenqibo@zj.chinamobile.com
}

\usepackage{bibentry}
\begin{document}

\maketitle

\begin{abstract}
Recent research on universal object detection aims to introduce language in a SoTA closed-set detector and then generalize the open-set concepts by constructing large-scale (text-region) datasets for training.
However, these methods face two main challenges: (i) how to efficiently use the prior information in the prompts to genericise objects and (ii) how to reduce alignment bias in the downstream tasks,  both leading to sub-optimal performance in some scenarios beyond pre-training.
To address these challenges, we propose a strong universal detection foundation model called CP-DETR, which is competitive in almost all scenarios, with only one pre-training weight.
Specifically, we design an efficient prompt visual hybrid encoder that enhances the information interaction between prompt and visual through scale-by-scale and multi-scale fusion modules.
Then, the hybrid encoder is facilitated to fully utilize the prompted information by prompt multi-label loss and auxiliary detection head.
In addition to text prompts, we have designed two practical concept prompt generation methods, visual prompt and optimized prompt, to extract abstract concepts through concrete visual examples and stably reduce alignment bias in downstream tasks.
With these effective designs, CP-DETR demonstrates superior universal detection performance in a broad spectrum of scenarios. For example, our Swin-T backbone model achieves 47.6 zero-shot AP on LVIS, and the Swin-L backbone model achieves 32.2 zero-shot AP on ODinW35.
Furthermore, our visual prompt generation method achieves 68.4 AP on COCO val by interactive detection, and the optimized prompt achieves 73.1 fully-shot AP on ODinW13.
\end{abstract}


\section{Introduction}
Universal object detection aims to detect objects of any category in any scene with one model weight.
The trend in research is to incorporate language modality, where textual descriptions of objects are encoded as text prompt vectors through language model~\cite{bert,clip}, and the classification results are represented by the similarity between the vectors and the image regions.
This flexible conceptual representation allows different object detection data to be trained jointly, aligning textual descriptions with visual representations.
Ultimately, in downstream tasks, universal object detection with zero-shot is achieved by modifying the textual descriptions of objects.

While using text prompts has been primarily favored in universal detection, they suffer from sub-optimal performance in downstream applications, where universal detectors fail to compete with specialist models in many scenarios and categories outside of pre-training.
A significant factor is the matching deficiency, where the detector produces mismatched results with the text description.
This deficiency arises from alignment mistakes between language and visual representations in pre-training, and there are both objective and subjective aspects to this bias.
Objectively, text descriptions follow a long-tailed pattern and different descriptions can refer to the same image region, so it is impractical to align all the texts and image regions accurately during pre-training.
Subjectively, it is difficult for users to accurately describe complex objects, such as specific mechanical devices, through language.
Most works~\cite{goldg,owlvit,owlvit2,detclipv3,glee} have been devoted to constructing larger pre-train datasets to address the alignment problems, but this requires significant costs.

Another factor is the paradigm of utilizing prompt information.
The work~\cite{glip} has shown that the early fusion paradigm performs significantly better than the late fusion paradigm after eliminating alignment bias through prompt tuning in the downstream tasks.
Late fusion paradigms~\cite{aaaizero} only use prompt vectors in the classification part, the location dependent on pre-training data distributions, which is poor in utilizing prompt information.
In contrast, the early fusion paradigm~\cite{groundingdino} has an additional cross-modal fusion phase.
It is easy to observe that the success of the early fusion paradigm lies in the cross-modal information interaction through fusion, where visual features are updated based on prompt information, and both classification and localization can be generalized in downstream scenes through prompt information.
Therefore, we believe that a key to improving the performance of universal detection lies in achieving effective cross-modal interaction between prompt and visual.

In this paper, our research is interested in constructing a strong universal detector that not only has superior zero-shot capability but also competes with specific models in all downstream tasks through a model weight.
For this, we propose CP-DETR, a model based on the early fusion paradigm that not only supports text prompts but also introduces visual prompts and optimized prompts to address alignment biases beyond pre-training.
Visual prompts avoid misalignment arising from subjective user description errors by providing visual examples to represent objects, e.g., by marking specific objects with boxes.
An optimized prompt provides a more direct solution by prompt tuning through downstream data annotation to align regions without changing the pre-training weights.
Interestingly, we note text prompts, visual prompts, and optimized prompts represent object concepts through high-dimensional vectors, so we use concept prompts to represent these vectors in a unified way and divide the whole model into two parts: detector and concept prompt generation.

The detector part determines the universal detection capability of the model, so we build the detector based on the SoTA DETR~\cite{dino} framework and exploit the prompting information through effective cross-modal interactions.
For effective cross-modal interaction, we design an efficient prompt visual hybrid encoder that updates visual and concept prompts via progressive single-scale fusion (PSF) and multi-scale fusion gating (MFG), avoiding confusion due to semantic gaps between different levels of visual features.
Due to DETR being a sparse detector framework, we added an auxiliary detection head and a prompt multi-label loss to facilitate the hybrid encoder to fully utilize different modal information in the interaction.

For the concept prompt generation part, CP-DETR supports both text prompts, visual prompts, and optimized prompts.
With text prompts, we use sentence-level representation to reduce computational overhead and encode them via CLIP~\cite{clip} encoder because of its better discriminability using larger-scale contrast learning.
For visual prompt, we design a visual prompt encoder that encodes the bbox as a query and adaptively aggregates concept representations from multi-scale features output by the visual backbone.
For optimized prompt, we design a super-class representation prompt tuning method to further improve the performance in downstream tasks by representing single categories through multiple vectors.

Through effective design, the CP-DETR demonstrates amazing universal detection capabilities, e.g., 
Using text prompt, it achieved a significant 32.2 zero-shot $AP$ on the ODinW35~\cite{odinw}. 
In the visual prompt interactive evaluation, it achieved 68.4 $AP$ on the COCO~\cite{coco} val.
Furthermore, using the optimized prompt method, it outperforms the previous SoTA model~\cite{glipv2} 5.1 average $AP$ on ODinW13~\cite{odinw} and can compete with full-model fine-tuned specialist models.

\section{Related Work}
\subsection{Text Prompted Universal Detection}
The recent work can be divided into early fusion and late fusion, depending on the degree of exploitation of the prompt. 
The late fusion-based method only utilizes the prompt information in the classification.
ViLD~\cite{vild}, RegionCLIP~\cite{regionclip} focuses on transferring knowledge from CLIP to detection.
The OWL-ViT~\cite{owlvit,owlvit2} and DetCLIP series~\cite{detclip,detclipv2,detclipv3} tend to directly align language and image regions through pre-training, therefore scaling up the data to 10$B$ and 50$M$ levels by pseudo-labeling, respectively.
The early fusion-based method considers the effect of the prompt on both classification and localization, using the prompt as a condition for image feature encoding.
GLIP~\cite{glip} fuses word-level text prompts with multi-scale image features through cross-attention and leverages grounding data to help learn aligned semantics.
Grounding DINO~\cite{groundingdino} further proposes language-guided query selection and cross-modality decoder to achieve denser fusion.
Then, APE~\cite{ape} and GLEE~\cite{glee} reduce the number of text prompts using sentence-level text encoding methods, significantly reducing the computational overhead of the fusion layer, and thus allowing more negative categories to be used during pre-training.
However, previous work uses all visual features to interact with prompts, ignoring the semantic gap of features at different levels in the backbone.
For this reason, we design a hybrid encoder to achieve efficient cross-modal interaction through progressive fusion from single to global scales.

\subsection{Visual Prompt}
Unlike text prompts, visual prompts use image information directly to refer to objects, avoiding misalignment due to incorrect descriptions.
Since late fused detectors have a double-tower structure, work~\cite{owlvit,ovdetr} adopts raw image as a visual prompt and leverages image-text-aligned representation to transfer the concept to a visual prompt.
MQ-Det~\cite{mqdet} uses a mixed representation of visual prompts and text prompts.
T-Rex2~\cite{trex2} uses visual instructions to achieve interactive detection, with input boxes and dots generating visual prompts to avoid context loss in cropped images.

\subsection{Optimized Prompt}
The optimized prompt is generated by prompt tuning, which has proved effective for alignment in the classification~\cite{coop}.
PromptDet~\cite{promptdet} uses this prompt as the context of the text prompt to guide the classification foundation model to achieve text and region alignment.
GLIP~\cite{glip} aligns concepts in downstream tasks by using optimized prompts as offsets to text prompts, noting that deep cross-modal fusion is critical to improving the effectiveness of prompt tuning.
Recent work~\cite{our} directly learning prompts avoids the dependence on text prompts and further improves performance.
The specificity of prompt tuning is that the optimization object is the activation value, which only reduces the alignment bias in the downstream task without changing the model.
Therefore, we believe that the evaluation metrics of the optimized prompt can better reflect detector universality.

\section{Method}

\begin{figure*}[t]
\centering
\includegraphics[width=0.8\textwidth]{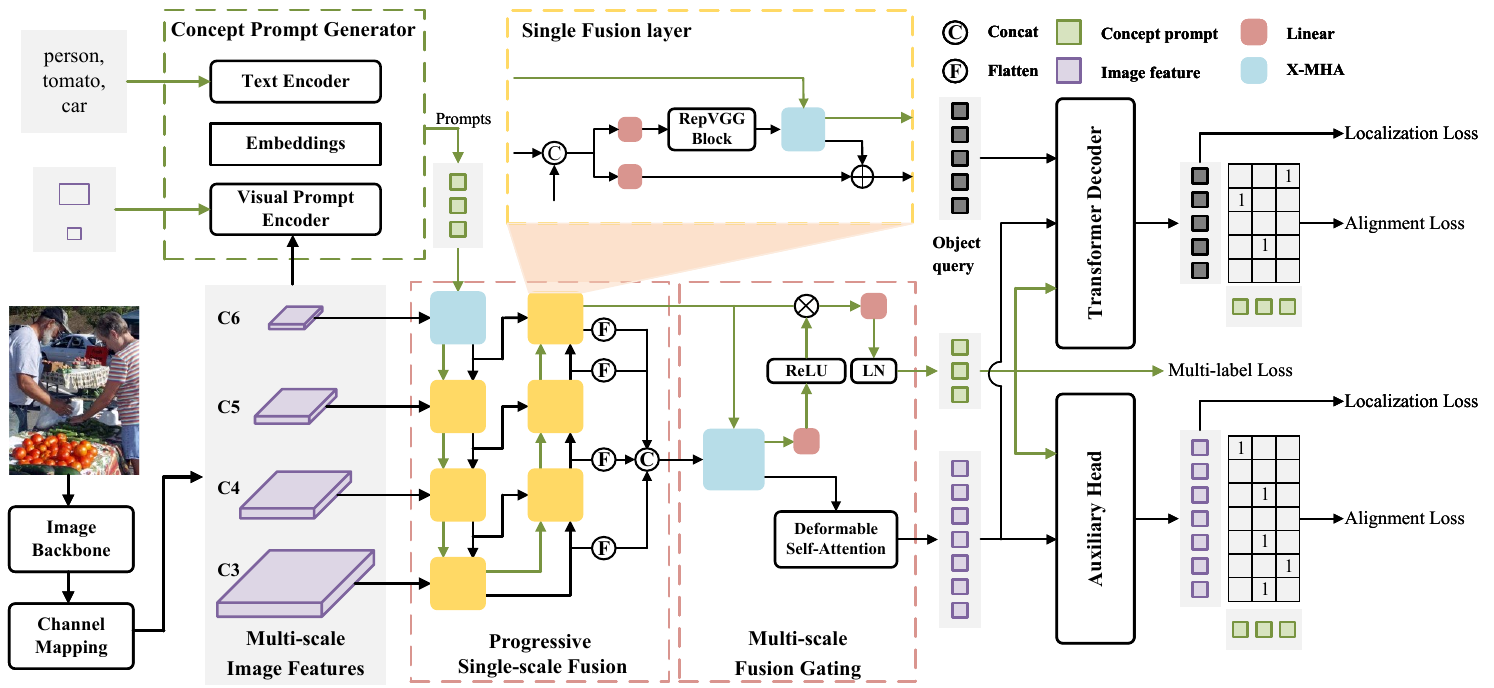} % Reduce the figure size so that it is slightly narrower than the column.
\caption{Overall architecture of CP-DETR.
First, the concept prompt generator (shown in green dashed box) encodes textual descriptions, referring boxes, or annotations as concept prompts. 
Then, the detector encodes the image as multi-scale feature maps and performs a cross-modal fusion of concepts and images using the proposed hybrid encoder (shown as the red dashed box). 
Finally, the transformer decoder predicts results.}
\label{method_fig1}
\end{figure*}

The overall architecture of the proposed CP-DETR is illustrated in figure \ref{method_fig1}, which consists of two parts: concept prompt generation and detection conditional on concept prompts.
We use concept prompt generators to encode different object references(e.g., text, box coordinates, etc.) into uniform vector space, which represent the object concepts and serve as conditional input detectors.
With different concept prompt generators, our model enables different workflows to handle alignment bias efficiently.

The detection part takes (prompts, image) pairs as input and outputs object boxes for the prompt's corresponding concepts.
For the image, the detector first obtains multi-scale image feature maps in 256 dimensions by image backbone and channel mapping. In this paper, we only use four scales: 1/8, 1/16, 1/32, and 1/64.
Then, a prompt visual hybrid encoder, which contains progressive single-scale fusion and multi-scale fusion gating, will be used for the mutual fusion of prompt and image features.
Following the previous work~\cite{groundingdino}, after obtaining fused features, 900 object queries are initialized language-guided query selection and updated by the 6-layer cross-modality decoder.
The training objectives for the transformer decoder are as follows:
\begin{equation}
    L_{decoder} = L_{localization} + L_{alignment}
    \label{eq0}
\end{equation}
where $L_{localization}$ contains GIoU~\cite{giou} loss and L1 loss, and $L_{alignment}$ is focal~\cite{qfocalloss} loss.

Due to the sparsity of object query, which could cause hybrid encoder sub-optimization, we introduce prompt multi-label classification loss and anchor-based auxiliary detection head in training as auxiliary supervision to facilitate cross-modal and cross-scale feature fusion.
The auxiliary supervision part will be removed during inference.

\subsection{Prompt Visual Hybrid Encoder}
Previous early fusion-based work~\cite{ape,glee,groundingdino,glip,glipv2} fused full-scale image feature maps and prompts simultaneously, which ignores the semantic gaps that exist between features at different scales.
However, due to the lack of semantic concepts and feature duplication, it is inefficient to perform cross-modal interaction on low-level feature maps in the early stages of fusion.
Therefore, we use a progressive single-scale fusion module that performs fusion scale-by-scale from high-level feature maps.
In order to avoid multi-scale information loss during scale-by-scale fusion, we also designed multi-scale fusion gating to enhance the fusion of critical information.
\subsubsection{Progressive Single-scale Fusion.}
The structure is illustrated in the left red dashed box of figure \ref{method_fig1}, which follows the top-down and bottom-up flow paths in~\cite{rtdetr,pan}.
The deepest $C6\in \mathcal{R}^{H/64 \times W/64 \times D}$ feature map has richer semantic concepts that help initially establish the connection between prompt and visual.
Therefore, we first use a cross-modality multi-head attention~\cite{glip}(X-MHA) to fuse $C6$ and prompt $P$ by:
\begin{equation}
    C6^{t=1},P^{l+1}= X\mbox{-}MHA(C6^{t=0},P^{l})
    \label{eq1}
\end{equation}
Where $l$ denotes the number of prompt fusions, $t\in {(0,1,2)}$ denotes the stage, and 0,1,2 denotes no fusion, top-down fusion, and bottom-up fusion, respectively.

Then, during top-down and bottom-up, we design a single fusion layer, as shown in the yellow dashed box of figure \ref{method_fig1}, with two neighboring scales of image features and prompts as inputs.
Specifically, neighboring feature maps are concatenated in the channel to obtain the hybrid feature $C_{ij}$, and the channels are adjusted through the linear layer and block~\cite{repvgg} to achieve cross-scale and implicit cross-modal information fusion simultaneously.
Then, using X-MHA to direct cross-modal fusion, obtains the updated prompt $P^{l+1}$ and image features $\bigtriangleup C$.
Finally, the image features $C_{j}^{t}$ of $j$ scale at stage $t$ are output by element-wise summation, which fuses $\bigtriangleup C$ with $C_{ij}$ after a linear layer.
The formula is as follows:
\begin{equation}
    \begin{split}
    & C_{ij} = concat(resize(C_{i}^{t} ),C_{j}^{t-1}) \\
    & P^{l+1}, \bigtriangleup C= X\mbox{-}MHA(Block(Linear(C_{ij})),P^{l}) \\
    & C_{j}^{t}  = \bigtriangleup C + Linear(C_{ij})
    \end{split}
    \label{eq2}
\end{equation}
\subsubsection{Multi-scales Fusion Gating.}
To avoid information loss due to scale-by-scale fusion processes, we propose to interact simultaneously at multi-scale feature maps.
The four-scale feature maps are flattened and then concatenated in the spatial dimension to form the full-scale feature $C_{all}$.
The fusion process of $C_{all}$ and prompt $P^{l}$ from PSF is as follows:
\begin{equation}
\begin{split}
    & P^{l+1}, C_{all}^{'} = X\mbox{-}MHA(C_{all},P^{l}) \\
    & P_{end}  = LN(Linear(ReLU(Linear(P^{l+1})*P^{l}))) \\
    & C_{all}^{''} = DeformAttn(C_{all}^{'})
\end{split}
\label{eq3}
\end{equation}

Where $DeformAttn$ is deformable self-attention~\cite{deformable}, $LN$ is Layernorm, $P_{end}$ denotes final concept prompts after full-scales information gating through dot product, $C_{all}^{''}$ denotes the image partial output of the hybrid encoder after full-scale image feature interaction by deformable self-attention.

\subsection{Auxiliary Supervision}
In the DETR architecture of detector training, both classification and location losses are implemented on the object queries.
However, due to the number of object query much smaller than the image features and using a one-to-one set matching scheme of label assignment, encoder output features get sparse supervision signals from the transformer decoder.
We argue that these sparse supervision signals will reduce the learning efficiency of cross-scale and cross-modal interactions in the hybrid encoder, leading to sub-optimal results.
Therefore, we introduce the auxiliary detection head and prompt multi-label loss to apply additional supervision to image features and conceptual prompts, respectively, which will facilitate fusion learning in the hybrid encoder.
\subsubsection{Auxiliary Detection Head.}
We choose an anchor-based detector head~\cite{atss} to facilitate training, which was shown effective in closed-set detection~\cite{codetr}.
The auxiliary head employs one-to-many label assignment and computes losses by anchors whose number is normal to image features, thus applying denser and direct supervision signals to image features.
We use a contrastive layer to replace the classification layer in the closed-set detector header and represent the category scores by the similarity $s_{mn}$ of prompt and image features as follows:

\begin{equation}
    s_{mn}=\frac{a^{m} \times Linear(P_{end}^{n})}{\sqrt{d} } + bias
    \label{eq4}
\end{equation}
Where $d$ is the number of feature channels, $bias$ is a learnable constant, $a^{m}$ denotes the image feature corresponding to the $m$-th anchor, and $P_{end}^{n}$ denotes the $n$-th concept vector.
With this simple modification, the closed-set detector head is converted to open-set form and thus can be used for auxiliary supervision in pre-training with class uncertainty.
The training objectives are as follows:
\begin{equation}
    L_{aux\_head} = L_{class}+L_{centerness}+L_{iou}
    \label{eq5}
\end{equation}
where $L_{class}$ is focal loss, $L_{centerness}$ is binary cross entropy loss, and $L_{iou}$ is GIoU loss.

\subsubsection{Prompt Multi-label Loss.}
In open-set pre-training, there are both positive prompts and a large number of negative prompts in each (image,prompts) pair, and the negative prompts don't have a corresponding object in the image.
Therefore, we could count the positivity and negativity of the prompts during the training process and automatically generate a multi-label annotation of $g$. 
The concept prompts output from the hybrid encoder are mapped to 1-dimensional through a $MLP$ layer, and the loss is computed as follows:
\begin{equation}
    L_{prompt}  = binary\_cross\_entropy(MLP(P_{end} ),g)
    \label{eq6}
\end{equation}

By applying a multi-label classification loss on a single modality, the concept prompts need learning to leverage the image information during the fusion process, thus rejecting the negative concept and retaining the positive prompts, making the fused concept prompts more discriminative.

\begin{figure}[t]
\centering
\includegraphics[width=0.8\columnwidth]{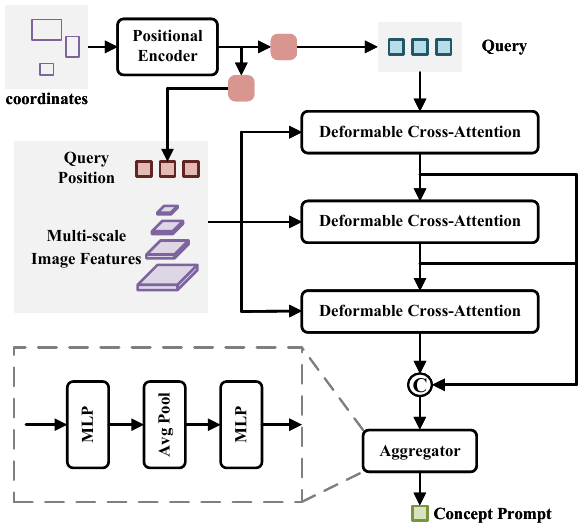} % 
\caption{The overall architecture of the visual prompt encoder. Coordinates of 2D boxes are encoded as query and query position vectors, and the concept prompt is aggregated from image features via three layers of deformable cross-attention.}
\label{method_fig2}
\end{figure}

\subsection{Concept Prompt Generator}
Text prompts successfully unify the training of different datasets and achieve zero-shot detection through a unified semantic space.
However, due to alignment bias, the detector is prone to associate with wrong objects when meeting long-tailed or inaccurately described text in downstream tasks.
For ODinW~\cite{odinw}, we observe that the performance of all existing universal models with zero-shot significantly lags behind the closed-set trained models.
Therefore, in order to reduce the impact of alignment bias on models in downstream tasks, CP-DETR also introduces two prompt generation methods, namely visual prompts and optimized prompts, to fully stimulate the universal detection capability of pre-trained models.

\subsubsection{Text Prompt.}
We select the pre-trained CLIP text encoder to extract text features and use average pooling to aggregate token-level text features into sentence-level concept prompt.
Only text prompts are used in CP-DETR pre-training, as this strategy was demonstrated efficiently in previous work~\cite{glip}.
In order to reduce the detection hallucination, which is predicting objects that are not present in the input image, we randomly sampled 80 categories or descriptions from the text dictionary as negative samples in the pre-training.
Unlike object detection datasets, Grounding and Referring Expression Comprehension (REC) datasets lack a unified category dictionary, so we construct a text dictionary online via a memory bank during training.
Then, in pre-training, the overall training objective is a linear combination of $L_{decoder}$, $L_{aux\_head}$, and $L_{prompt}$.

\subsubsection{Visual Prompt.}
Figure \ref{method_fig2} shows encoder structure, where $N$ normalized box coordinates are encoded by sine-cosine position encoder to obtain vector $r\in \mathcal{R}^{N \times 128}$, which respectively through two linear layers generates query embedding $q$ and query position embedding $q_{pos}$.
Then, concept information is extracted from the image features by cross-attention, with $q_{pos}$ limiting the extraction range to ensure information is relevant to box content.
We use three layers of attention and concatenate the output query of each layer by channel, and finally generate the concept prompt for the corresponding category through an aggregator, which is shown in the dashed box in figure \ref{method_fig2}.
When training the encoder, we freeze the pre-training weights and use the box sampling method of work~\cite{trex2}.
Since we train the encoder after pre-training, where text prompts can be considered as concept prompt ground truths on the pre-training data, in addition to $L_{decoder}$, we also use MSE loss for direct supervision, with the overall training objective as follows:
\begin{equation}
    L_{visual\_prompt}=\frac{1}{K}\sum_{i=0}^{K}(P_{v}^{i}-P_{t}^{i})^{2}  +L_{decoder}
    \label{eq7}
\end{equation}
where $K$ is the number of positive categories, $P_{v}$ denotes the concept prompt obtained by a visual prompt encoder, and $P_{t}$ denotes the concept prompt obtained by the text encoder.

\subsubsection{Optimized Prompt.}
We freeze all model parameters and initiate concept prompts with learnable embedding layers, which are fine-tuned to get aligned concept prompts.
In addition, we propose the super-class representation considering the case where different classes may be labeled as the same class in downstream scenarios.
Specifically, class $I$ corresponds to $M$ prompts, and the correspondence is saved through a mapping table.
Finally, the maximum similarity value was extracted from the $M$ prompts as the classification score.
Since hybrid encoder optimization is not required, the training objective contains only $L_{decoder}$.

\section{Experiments}
\subsection{Training Datasets.}
We use multiple datasets with region-text annotations from different sources for joint training.
For the object level, we use publicly available detection datasets, which contain Objects365~\cite{o365} (O365), OpenImages\cite{openimage} (OI), V3Det~\cite{v3det}, LVIS~\cite{lvis} and COCO~\cite{coco} datasets.
For grounding or REC data, we used the GoldG~\cite{goldg}, RefCOCO/+/g~\cite{refcoco,refcocog}, Visual Genome~\cite{vg} (VG) and PhraseCut~\cite{phrasecut} datasets, with a memory bank set length of 1000 in pre-training.
where GoldG, RefCOCO/+/g, we used the cleaned labels from GLIP~\cite{glip} and we combined RefCOCO/+/g into RefC by removing duplicate samples.
For GoldG, PhraseCut, and VG, where object phrases are treated as categories.
For RefC, we treat the entire description as a category.
It is worth noting that the training labels we use all come from publicly available datasets and do not scale up the data by pseudo-labeling image-text pair data as most work~\cite{detclipv3,glee} does.
\subsection{Implementation Details.}
In our experiments, we developed two model variants, CP-DETR-T and CP-DETR-L, by using Swin-Tiny and Swin-Large~\cite{swin} as image backbone, respectively.
We used CLIP-L~\cite{eva} as the text encoder in all variants and only fine-tuned it during pre-training.
For CP-DETR-T, we use O365, V3Det, and GoldG for pre-training with a total training epoch of 30.
For CP-DETR-L, we train 1$M$ iterations using all training datasets.
In all experiments, we use AdamW as the optimizer with weight decay set to 1e-4 and set a minibatch to 32 on 8 A100 40GB GPUs.
In pre-training, the learning rate was set to 1e-5 for the text encoder and image backbone and 1e-4 for the rest of the modules, and a decay of 0.1 was applied at 80\% and 90\% of the total training steps.
In visual prompt training, the O365, V3Det, GoldG, and OI datasets are used, the learning rate of the visual prompt encoder is set to 1e-4, and the training is performed for 0.5$M$ iterations.
In the optimized prompt, the learning rate of the embedding layer is set to 5e-2, the total number of training epochs is 24, and a decay of 0.1 is applied at 80\% of the total training steps.
\subsection{Evaluation Benchmark.}

\begin{table*}[t]
\centering
\begin{small}
\begin{tabular}{l|l|cc|cc|c|c}
\hline
\multirow{2}{*}{Method} & \multirow{2}{*}{Backbone} & \multicolumn{2}{c|}{COCO} & \multicolumn{2}{c|}{LVIS} & RefC            & ODinW35 \\ \cline{3-8} 
                        &                           & val       & test-dev      & minival       & val       & refcoco/+/g    & test    \\ \hline
GLIP-T~\cite{glip}                  & Swin-T                    & 46.3      & -             & 26.0          & 17.2      & 50.4/49.5/66.1 & 19.6    \\
Grounding-DINO-T~\cite{groundingdino}        & Swin-T                    & 48.4      & -             & 27.4          & 20.1      & 50.8/51.6/60.4 & 22.3    \\
YOLO-World-L~\cite{yoloworld}            & YOLOv8-L                  & 45.1      & -             & 35.4          & -         & -              & -       \\
DetCLIPv3-T~\cite{detclipv3}             & Swin-T                    & 47.2      & -             & 47.0          & 38.9      & -              & -       \\
T-Rex2-T~\cite{trex2}                & Swin-T                    & 45.8      & -             & 42.8          & 34.8      & -              & 18.0    \\
CP-DETR-T               & Swin-T                    & 52.0      & 52.2             & 47.6          & 39.9         & 43.7/42.2/52.6          & 27.3       \\ \hline
GLIPv2-H~\cite{glipv2}                & Swin-H                    & -         & \textcolor{grey}{60.6}          & \textcolor{grey}{59.8}          & -         & -              & -       \\
Grounding-DINO-L~\cite{groundingdino}        & Swin-L                    & \textcolor{grey}{60.7}      & -             & 33.9          & -         & \textcolor{grey}{90.6}/\textcolor{grey}{82.8}/\textcolor{grey}{86.1} & 26.1    \\
OmDet-Turbo-B~\cite{omdet2}           & ConvNeXt-B                & 53.4      & -             & 34.7          & -         & -              & 30.1    \\
T-Rex2-L~\cite{trex2}               & Swin-L                    & 52.2         & -             & 54.9          & 45.8      & -              & 22.0       \\
OWL-ST~\cite{owlvit2}                  & CLIP L/14                 & -         & -             & 40.9          & 35.2      & -              & -    \\
UNINEXT-H~\cite{uninext}               & ViT-H                     & \textcolor{grey}{60.6}      & -             & 18.3          & 14.0      & \textcolor{grey}{92.6}/\textcolor{grey}{85.2}/\textcolor{grey}{88.7} & -       \\
DetCLIPv2-L~\cite{detclipv2}             & Swin-L                    & -         & -             & 44.7          & 36.6      & -              & -       \\
DetCLIPv3-L~\cite{detclipv3}             & Swin-L                    & 48.5      & -             & 48.8          & 41.4      & -              & -       \\
GLEE-Pro~\cite{glee}                & ViT-L                     & \textcolor{grey}{62.0}      & \textcolor{grey}{62.3}          & -             & \textcolor{grey}{55.7}      & \textcolor{grey}{91.0}/\textcolor{grey}{82.6}/\textcolor{grey}{86.4} & -       \\
APE(D)~\cite{ape}                  & ViT-L                     & \textcolor{grey}{58.3}      & -             & \textcolor{grey}{64.7}          & \textcolor{grey}{59.6}      & \textcolor{grey}{84.6}/\textcolor{grey}{76.4}/\textcolor{grey}{80.0} & 28.8    \\
CP-DETR-L               & Swin-L                    & \textcolor{grey}{62.8}      & \textcolor{grey}{62.7}          & \textcolor{grey}{65.9}          & \textcolor{grey}{60.3}      & \textcolor{grey}{90.7}/\textcolor{grey}{81.4}/\textcolor{grey}{85.6} & 32.2    \\ \hline
\end{tabular}
\end{small}
\caption{Comparison with state-of-the-art universal models on multiple datasets through text prompts.
Black numbers indicate zero-shot.
Gray numbers indicate that the model pre-training contains the training parts of this dataset.}
\label{et1}
\end{table*}

\begin{table*}[t]
\centering
\begin{small}
\setlength{\tabcolsep}{1mm}{
\begin{tabular}{lc|cccccccccccccc}
\hline
\rot{Method} &\rot{Tune}   &\rot{PascalVOC} &\rot{AerialDrone} &\rot{Aquarium} &\rot{Rabbits} &\rot{EgoHands} &\rot{Mushrooms} &\rot{Packages} &\rot{Raccoon} &\rot{Shellfish} &\rot{Vehicles} &\rot{Pistols} &\rot{Pothole} &\rot{Thermal} &\rot{\textbf{Average}} \\ \hline
GLEE-Pro~\cite{glee}         & full   & 72.6      & 36.5        & 58.1     & 80.5    & 74.1     & 92.0      & 67.0     & 76.5    & 66.4      & 70.5     & 66.4    & 55.7    & 80.6    & 69.0    \\
GLIP-L~\cite{glip}           & full   & 69.6      & 32.6        & 56.6     & 76.4    & 79.4     & 88.1      & 67.1     & 69.4    & 65.8      & 71.6     & 75.7    & 60.3    & 83.1    & 68.9    \\
GLIPv2-H~\cite{glipv2}         & full   & 74.4      & 36.3        & 58.7     & 77.1    & 79.3     & 88.1      & 74.3     & 73.1    & \textbf{70.0}      & 72.2     & 72.5    & 58.3    & 81.4    & 70.4    \\
OmDet-B~\cite{omdet1}          & full   & 71.2      & 27.5        & 52.7     & 76.5    & 77.4     & \textbf{93.6}      & 73.7     & 74.3    & 57.7      & 64.5     & 74.2    & 56.9    & 83.3    & 68.0    \\
DetCLIPv2-L~\cite{detclipv2}      & full   & 74.4      & 44.1        & 54.7     & \textbf{80.9}    & 79.9     & 90        & 74.1     & 69.4    & 61.2      & 68.1     & 80.3    & 57.1    & 81.1    & 70.4    \\
DetCLIPv3-L~\cite{detclipv3}      & full   & 76.4      & \textbf{51.2}        & 57.5     & 79.9    & \textbf{80.2}     & 90.4      & 75.1     & 70.9    & 63.6      & 69.8     & \textbf{82.7}    & 56.2    & \textbf{83.8}    & 72.1    \\
GLIP-L~\cite{glip}           & prompt & 72.9      & 23.0        & 51.8     & 72.0    & 75.8     & 88.1      & 75.2     & 69.5    & 73.6      & 72.1     & 73.7    & 53.5    & 81.4    & 67.9    \\
GLIPv2-H~\cite{glipv2}         & prompt & 71.2      & 31.1        & 57.1     & 75.0    & 79.8     & 88.1      & 68.6     & 68.3    & 59.6      & 70.9     & 73.6    & \textbf{61.4}    & 78.6    & 68.0    \\
Grounding-DINO-T~\cite{our} & prompt & 71.7      & 34.2        & 53.0     & 75.8    & 73.4     & 88.1      & 75.6     & 74.3    & 58.7      & 68.0     & 73.6    & 52.3    & 81.5    & 67.7    \\
CP-DETR-T        & prompt & 74.2         & 37.7           & 54.4        & 78.4       & 75.5        & 88.1         & 72.0        & 72.8       & 61.0         & 72.9        & 75.9       & 54.4       & 82.2       & 69.2       \\
CP-DETR-L        & prompt & \textbf{80.5}      & 47.9        & \textbf{60.3}     & 77.5    & 79.0     & 90.4      & \textbf{76.4}     & \textbf{77.0}    & 68.9      & \textbf{73.4}     & 81.5    & 55.9    & 81.5    & \textbf{73.1}    \\ \hline
\end{tabular}}
\end{small}
\caption{Comparison with state-of-the-art universal models on multiple datasets through fine-tuning.
A tune of type full indicates fine-tuning the full model.
A tune of type prompt indicates optimizing prompt only.}
\label{et3}
\end{table*}

\begin{table}[t]
\centering
\setlength{\tabcolsep}{1mm}{
%\resizebox{.95\columnwidth}{!}{
\begin{small}
\begin{tabular}{lc|ccc}
\hline
Method    & Backbone & COCO-val & LVIS-minival & ODinW35 \\ \hline
T-Rex2-T  & Swin-T   & 56.6     & 59.3         & 37.7    \\
T-Rex2-L  & Swin-L   & 58.5     & 62.5         & 39.7    \\
CP-DETR-T & Swin-T   & 61.8        & 64.1            & 41.0       \\
CP-DETR-L & Swin-L   & 68.4        & 71.6            & 50.6       \\ \hline
\end{tabular}
\end{small}}
\caption{Comparison with universal models on multiple datasets through interactive object detection.}
\label{et2}
\end{table}

We evaluated the universal detection ability on the COCO, LVIS, ODinW~\cite{odinw} and RefCOCO/+/g benchmarks.
ODinW contains 35 real-world scenarios that can reflect the model's universality in downstream tasks.
For COCO, LVIS, and ODinW, the $AP$ is an evaluation metric.
Following work~\cite{groundingdino}, we also used RefCOCO/+/g to evaluate the ability of the model to understand complex textual descriptions with the P@0.5 metric.

\subsection{Comparison with Universal Detectors}
By switching among the three concept prompt generation methods, we demonstrate the universality and effectiveness of CP-DETR as an object detection model, both in the pre-training domain and downstream scenarios, while ensuring state-of-the-art performance.
In all evaluations, CP-DETR only uses one weight.
\subsubsection{Text Prompt Direct Evaluation.}
In this evaluation, we use all category names or description sentences of the benchmark as text prompt inputs, consistent with previous work~\cite{ape} settings.
Depending on whether the benchmark's training set is used in pre-training, the text prompt-based evaluation can be categorized into zero-shot and full-shot.
We primarily use CP-DETR-T to evaluate the effectiveness of our method on zero-shot.
As shown in table \ref{et1}, CP-DETR-T outperforms all similarly sized previous models in COCO and LVIS benchmarking, with +3.6$AP$ and +20.2$AP$ compared to baseline Grounding DINO.
The method closest to ours in terms of zero-shot performance is DetCLIPv3-T, which not only uses 1.61$M$ of O365, V3Det, and GoldG as we do, but also an extra 50$M$ of private data GranuCap50M, which indicates that our method is sufficiently effective in terms of concept generalization.
CP-DETR-T has limitations on RefC, which we believe are due to the pre-training containing only object phrases and lacking the descriptive sentences required in the RefC evaluation.
On the ODinW35 benchmark, we observed that several datasets showed significant quality issues in terms of annotated category names, so we followed the APE~\cite{ape} evaluation setup, and our CP-DETR-L set a new zero-shot record with an average of 32.2$AP$ across 35 datasets.

A universal model should have concept generalization capabilities and perform well in scenarios that have already been seen in pre-training.
Due to CP-DETR-L's pre-training data containing COCO, LVIS, and RefC, we use it for full-shot comparisons with other state-of-the-art universal models.
As shown at the bottom of table \ref{et1}, the CP-DETR-L simultaneously achieves state-of-the-art performance or competitive performance in all object detection benchmarks, with +2.1$AP$ in COCO-val compared to baseline~\cite{groundingdino}.
On the RefC benchmark, CP-DETR achieved comparable results to Grounding DINO-L, showing that the sentence feature as a concept prompt is sufficient to represent complex textual descriptions.
Notably for the COCO and LVIS parts of the evaluation, the state-of-the-art APE~\cite{ape} and GLEE~\cite{glee} performed well on only one of them, even though they used a larger backbone and stronger large-scale jittering data augmentation methods.
And CP-DETR performs well on both benchmarks, proving that our method remembers and distinguishes all seen concepts well.

\subsubsection{Fine-tuning Evaluation.}
Table \ref{et3} shows the comparison results with the state-of-the-art universal detection models on 13 subsets in ODinW, which are fine-tuned using prompt or full model.
Optimized prompts reduce alignment bias by adjusting concept prompts and can truly reflect the universality of the detector.
The significant performance advantage of our approach in this setting, along with the +5.1$AP$ compared to GLIPv2-H~\cite{glipv2} in terms of average metrics, demonstrates the strong generalization of CP-DETR in downstream scenarios, and we believe that this advantage stems from our design, which better facilitates the use of prompt information.
Even compared to the approach of applying full model fine-tuning, CP-DETR-L still achieved state-of-the-art or competitive performance in 13 subsets with only optimized prompts, and set a new record of 73.1$AP$ on average.
This phenomenon indicates that CP-DETR can achieve competitive performance with a specific model in downstream scenarios by using a pre-trained weight, greatly enhancing the application value of the universal model in the real world.

\subsubsection{Visual Prompt Interactive Evaluation.}
Since the concept prompt generation of visual prompt requires boxes as input, we use interactive evaluation, unlike the interactive process of T-Rex2~\cite{trex2}, we avoid introducing category priors.
For the test image with $M$ total dataset categories and $N$ positive categories, we randomly chose a GT box as the visual prompt input for the positive category and used text prompts for the remaining $M-N$ negative categories.
As shown in table \ref{et2}, our method significantly outperforms previous work~\cite{trex2} in all benchmarks. 
CP-DETR is the first to implement interactive detection based on visual prompts in the early fusion paradigm, which is more effective in exploiting prompt information than the late fusion paradigm~\cite{trex2}.
Comparing table \ref{et1} and \ref{et2}, it can be observed that visual prompts outperform text prompts, with +18.4$AP$ on ODinW35, indicating that visual prompts can reduce alignment bias and have a strong application in interactive scenarios such as auxiliary labelling.

\subsection{Ablation}
\begin{table}[t]
\centering
\setlength{\tabcolsep}{1mm}{
\begin{small}
\begin{tabular}{c|l|c|c}
\hline
\multirow{2}{*}{Row} & \multirow{2}{*}{Model Set}   & LVIS minival & ODinW13   \\ \cline{3-4} 
                    &                              & Zero-shot    & Full-shot \\ \hline
0                   & CP-DETR(base model)          & 44.3         & 64.0         \\
1                   & replaced by DINO encoder   & 42.2            & 58.5         \\
2                   & w/o MFG                      & 42.8         & 62.3         \\
3                   & add prompt multi-label loss & 44.8         & 64.2         \\
4                   & add row3 and auxiliary head & 44.7         & 64.9         \\
5                   & add row3, row4 and super-class              & 44.7         & 67.0         \\ \hline
\end{tabular}
\end{small}}
\caption{Ablations for our model with a Swin-T backbone. 
The full shot is achieved by the optimized prompt.}
\label{et4}
\end{table}
In this section, we conducted ablation experiments.
The different variants models all use the Swin-T backbone and are trained on O365, V3Det, and GoldG for 12 epochs.
In order to show the impact of various components on the universality of the detector, we not only performed a zero-shot evaluation on LVIS but also employed a full-shot optimized prompt on ODinW13.

Table \ref{et4} demonstrates the effectiveness of the different designs, where row 0 is the CP-DETR base model without the inclusion of auxiliary supervision and super-class representation.
The prompt visual hybrid encoder was ablated in rows 1 and 2; the results show that the hybrid fusion approach of PSF and MSG reduces the difficulty of alignment and contributes to the zero-shot generalization of concepts, and the metric is improved by 0.6$AP$ and 1.5$AP$ on LVIS, respectively.
The hybrid encoder is the most important improvement, with the ODinW13 full-shot metric upgraded from 58.5$AP$ in row 1 to 64.0$AP$.
This improvement reveals the importance of effective cross-modal fusion for universal location, encouraging the model to decode object boxes based on information in the prompt.
Rows 3 and 4 experiments show that auxiliary supervision facilitates the hybrid encoder in learning cross-modal knowledge during the training phase and modestly improves the zero-shot and full-shot metrics performance.
Row 5 further adds super-class in prompt fine-tuning, with the number set to 10 in the experiment, i.e., 10 prompt vectors represent a category.
This design can effectively handle situations where different objects in downstream scenes are represented in the same category, thus further improving the ODinW13 metric to 67.0$AP$.
See the Technical Appendix for more results.

\section{Conclusion}
In this paper, we propose a universal detector CP-DETR that enables prompt-conditional detection through efficient prompt visual fusion.
We focus on downstream applications and achieve SoTA zero-shot performance through text prompts.
Furthermore, with visual prompt and optimized prompt, CP-DETR with only one weight can compete with the full model fine-tuned methods in downstream scenarios, demonstrating its superior universality.

\bibliography{aaai25}

\section{Appendix}
\subsection{More Implementation Details}
For Hungarian matching, we following previous works~\cite{groundingdino,ape}, and set the weight of alignment costs, L1 costs, and GIoU costs as 2.0, 5.0, and 2.0, respectively. 
The corresponding loss weights in the transformer decoder are 1.0, 5.0 and 2.0, respectively.
Since the transformer decoder computes losses at each layer, to balance the contribution of different losses, we refer to previous work~\cite{codetr} and set the prompt multi-label loss weight to 6, and the class loss, centerness loss and IoU loss in the auxiliary detection header to 6, 6, and 12, respectively.

Due to GPUs resource limitation and in order to reduce memory spikes, we apply gradient checkpoints and automatically mixed precision (AMP) techniques in the images backbone and prompt visual hybrid encoder.
Both our CP-DETR-T and CP-DETR-L use 4 scales image features, where 1/8 to 1/32 is from the image backbone and 1/64 is from channel mapping downsampling.
For images augmentation, we use the default DETR~\cite{dino} augmentation in MMDetection~\cite{mmdetection} toolbox, which includes multi-scale training and random flip.

\subsection{More Training Data Details}
We compare the data usage of CP-DETR with other methods in table \ref{appendet1}.
It can be found that most of the methods construct private training annotations to better align different modalities by extending the richness of the training samples.
In contrast, CP-DETR achieves excellent zero-shot performance using only publicly available annotations. We believe there are two reasons for this: firstly, the CLIP~\cite{eva} text encoder has seen ample visual concepts and the proposed design is effective enough in exploiting concept information. Second, our using sentence-level representations, where a large number of negative categories can be used in a batch, reduces the illusion.

In addition, the sampling rates we configured for the different datasets are shown in table \ref{appendet2}.
It should be noted that GoldG~\cite{goldg} data contains both GQA and Flickr30k components. However, we found that multiple samples in GQA shared a single image, so we merged these samples and reduced the data size from 0.62$M$ to 0.09$M$.
O365 contains v1 and v2 versions, based on previous studies~\cite{groundingdino,glip}, we use v1 on CP-DETR-T and v2 on CP-DETR-L.

\begin{table*}[t]
\centering
\setlength{\tabcolsep}{1mm}{
\begin{small}
\begin{tabular}{l|l|l|l}
\hline
Method           & Backbone  & Publicly Available Data            & Private Annotated Data \\ \hline
GLIP-T~\cite{glip}           & Swin-T    & O365,GoldG                                      & Cap4M                 \\
Grounding-DINO-T~\cite{groundingdino} & Swin-T    & O365,GoldG                                      & Cap4M                 \\
YOLO-World-L~\cite{yoloworld}     & YOLOv8-L  & O365,GoldG                                      & CC3M                  \\
DetCLIPv3-T~\cite{detclipv3}      & Swin-T    & O365,V3Det,GoldG                                & GranuCap50M           \\
T-Rex2-T~\cite{trex2}         & Swin-T    & O365,OI,GoldG,HierText,CrowdHuman               & CC3M,SBU,LAION        \\
CP-DETR-T        & Swin-T    & O365,V3Det,GoldG                                & -                     \\ \hline
GLIPv2-H~\cite{glipv2}         & Swin-H    & O365,OI,VG,ImageNetBoxes,COCO,GoldG                              & CC15M,SBU             \\
Grounding-DINO-L~\cite{groundingdino} & Swin-L    & O365,OI,GoldG,COCO,RefC                         & Cap4M                 \\
UNINEXT-H~\cite{uninext}        & ViT-H     & O365,COCO,RefC,SOT\&VOS,MOT\&VIS,RVOS          & -                     \\
OWL-ST~\cite{owlvit2}           & CLIP L/14 & -                                               & WebLI2B               \\
T-Rex2-L~\cite{trex2}         & Swin-L    & O365,OI,GoldG,HierText,CrowdHuman               & CC3M,SBU,LAION        \\
DetCLIPv3-L~\cite{detclipv3}      & Swin-L    & O365,V3Det,GoldG                                & GranuCap50M           \\
GLEE-Pro~\cite{glee}         & ViT-L     & O365,VG,COCO,OI,LVIS,BDD,RefC,RVOS,VIS        & -                 \\
APE(D)~\cite{ape}           & ViT-L     & COCO,LVIS,O365,OI,VG,RefC,GoldG,PhraseCut       & SA-1B                 \\
CP-DETR-L        & Swin-L    & O365,V3Det,GoldG,OI,VG,RefC,COCO,LVIS,PhraseCut & -                     \\ \hline
\end{tabular}
\end{small}}
\caption{A detailed list of training data for different models. VIS consists of YTVIS19, YTVIS21, and OVIS. GoldG consists of GQA and Flickr30k. Private annotated data, indicating that the annotation of the corresponding data is privately constructed by them and is not publicly available.}
\label{appendet1}
\end{table*}

\begin{table*}[t]
\centering
\begin{small}
\begin{tabular}{l|l|c|c|cc|c|c|c|c|c|c}
\hline
\multirow{2}{*}{Model}     & \multirow{2}{*}{Target} & \multirow{2}{*}{O365} & \multirow{2}{*}{V3Det} & \multicolumn{2}{c|}{GoldG}                                & \multirow{2}{*}{OI} & \multirow{2}{*}{VG} & \multirow{2}{*}{RefC} & \multirow{2}{*}{COCO} & \multirow{2}{*}{LVIS} & \multirow{2}{*}{PhraseCut} \\ \cline{5-6}
                           &                         &                       &                        & \multicolumn{1}{l|}{GQA} & \multicolumn{1}{l|}{Flickr30k} &                     &                     &                       &                       &                       &                            \\ \hline
\multirow{2}{*}{CP-DETR-T} & Pre-training            & 1                     & 1                      & \multicolumn{1}{c|}{3}   & 1                              & -                   & -                   & -                     & -                     & -                     & -                          \\
                           & Visual Prompt           & 1                     & 1                      & \multicolumn{1}{c|}{1}   & 1                              & 1                   & -                   & -                     & -                     & -                     & -                          \\ \hline
\multirow{2}{*}{CP-DETR-L} & Pre-training            & 1                     & 1                      & \multicolumn{1}{c|}{3}   & 1                              & 1                   & 2                   & 3                     & 2                     & 2                     & 1                          \\
                           & Visual Prompt           & 1                     & 1                      & \multicolumn{1}{c|}{1}   & 1                              & 1                   & -                   & -                     & -                     & -                     & -                          \\ \hline
\end{tabular}
\end{small}
\caption{Training data sampling ratio configures.}
\label{appendet2}
\end{table*}

\begin{table}[!ht]
\centering
\setlength{\tabcolsep}{1mm}{
\begin{small}
\begin{tabular}{l|cc|cc}
\hline
\multirow{2}{*}{Model} & \multicolumn{2}{c|}{FPS(bs=1)}                                  & \multicolumn{2}{c}{FPS(bs=4)}                                  \\ \cline{2-5} 
                       & \multicolumn{1}{l}{1 classes} & \multicolumn{1}{l|}{80 classes} & \multicolumn{1}{l}{1 classes} & \multicolumn{1}{l}{80 classes} \\ \hline
Grounding-DINO-T       & 9.2                           & 5.0                             & 9.3                           & 5.7                            \\
CP-DETR-T              & 12.2                          & 11.2                            & 14.9                          & 13.3                           \\ \hline
Grounding-DINO-L       & 3.0                           & 2.0                             & 2.7                           & 1.8                            \\
CP-DETR-L              & 5.5                           & 5.4                             & 5.2                           & 4.9                            \\ \hline
\end{tabular}
\end{small}}
\caption{Comparison results of model inference efficiency. The $bs$ denotes the size of the batchsize used for single inference. FPS indicates the number of images processed by the model per second, and larger indicates more efficient inference.}
\label{appendet3}
\end{table}

\subsection{Additional Experiment}

\begin{figure}[t]
\centering
\includegraphics[width=0.9\columnwidth]{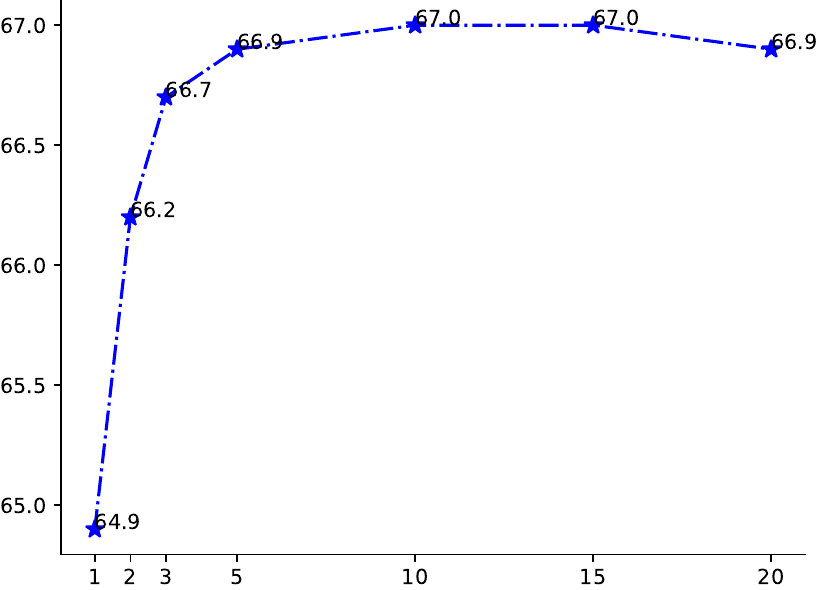} % 
\caption{Ablation results for the super-class representation length of optimized prompt in CP-DETR-T.}
\label{append1}
\end{figure}

\begin{figure*}[!pt]
\centering
\includegraphics[width=2\columnwidth]{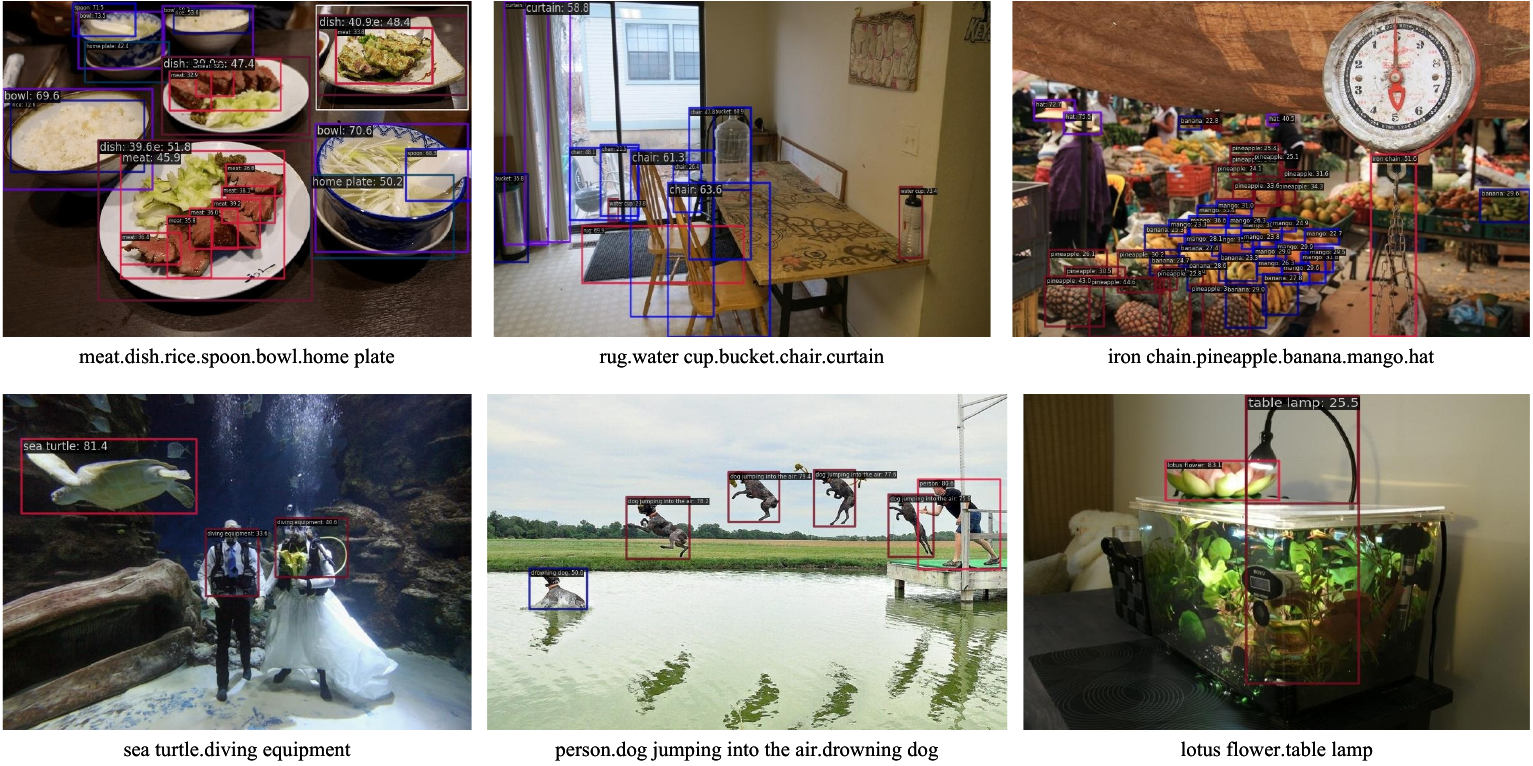} % 
\caption{Visualizations of CP-DETR-L zero-shot outputs.}
\label{append2}
\end{figure*}

\begin{table*}[t]
\centering
\begin{tabular}{lcccccccccc}
\bottomrule
\multirow{2}{*}{Method} & \multirow{2}{*}{Data scale} & COCO-val & \multicolumn{4}{c}{LVIS-minival} & \multicolumn{4}{c}{LVIS-val} \\
                        &                             & AP$_{all}$    & AP$_{all}$   & AP$_{r}$    & AP$_{c}$   & AP$_{f}$   & AP$_{all}$  & AP$_{r}$   & AP$_{c}$  & AP$_{f}$  \\ \midrule
\multicolumn{11}{c}{Open-source}                                                                                                   \\ \midrule
Current SOTA in each item            & N/A                         & 53.4     & 43.4    & 34.5   & 41.2  & 46.9  & 34.7   & 26.9  & 32.0 & 41.3 \\ \midrule
\multicolumn{11}{c}{Closed-source}                                                                                                 \\ \midrule
DetCLIPv3-L             & 50M                         & 48.5     & 48.8    & 49.9   & 49.7  & 47.8  & 41.4   & 41.4  & 40.5 & 42.3 \\
Trex-2-L                & 6.5M                        & 52.2     & 54.9    & 49.2   & 54.8  & 56.1  & 45.8   & 42.7  & 43.2 & 50.2 \\
Grounding DINOv1.5 Pro  & 20M                         & 54.3     & 55.7    & 56.1   & 57.5  & 54.1  & 47.6   & 44.6  & 47.9 & 48.7 \\
Grounding DINOv1.6 Pro  & 30M                         & 55.4     & 57.7    & 57.5   & 60.5  & 55.3  & 51.1   & 51.5  & 52.0 & 50.1 \\ \midrule
CP-DETR-Pro             & 1.1M                        & 55.4     & 58.2    & 60.6   & 59.2  & 56.8  & 51.6   & 51.3  & 51.6 & 51.8 \\ \bottomrule
\end{tabular}
\caption{Zero-shot performance of CP-DETR-Pro on the COCO, LVIS-minival and LVIS-val benchmarks compared to previous methods.}
\label{appendet4}
\end{table*}

Since ablation experiments in the main manuscript reveal that the super-class representation has a large performance gain for optimized prompts, we experimented with the super-class representation length as well.
As shown in figure~\ref{append1}, the performance on the downstream task gradually improves as the representation length increases, approaching saturation at 10, so we use 10 as the default length for optimized prompts.

In addition, table~\ref{appendet3} compares the model size and inference efficiency of CP-DETR and Grounding DINO~\cite{groundingdino}. 
For a fair comparison, we use the Grounding DINO implemented in MMDetection~\cite{mmdetection}.
Automatically mixed precision was kept off in all tests.
The results show that our model is more inference efficient.
There are three main reasons for this, firstly our cross-modal interactions are scale-by-scale, which has less computational overhead compared to works~\cite{groundingdino} which interact at all scales. Second, we rely on PAN~\cite{pan} structure to fuse image features instead of dense deformable self-attention~\cite{deformable} operation.
Finally, Grounding DINO-L uses 1/4 to 1/64 of the image feature maps, while we only use 1/8 to 1/64 of the image feature maps on the largest scale model, requiring fewer image features to be processed.

\subsection{Limitation}
Although our model exhibits strong universal detection performance, it still has some challenges. 
On the one hand, the pre-training of CP-DETR relies heavily on text quality, yet there are potential descriptive conflicts between different datasets, e.g., the noun "mouse", which denotes a computer device in most of the data, whereas it is used to describe an animal in some scenarios.
We believe that such textual defects will reduce the model's optimisation efficiency and affect the zero-shot capability.
On the other hand, since we use average pooling to obtain sentence-level text prompts, this may lead to incorrect optimisation of objects in sentences.
For example, if the training text "person wearing helmet" exists, the zero-shot of "helmet" will most likely frame out the person with a helmet after pre-training, assuming that there is a lack of category annotation of "helmet" in the data.
In addition, it can be observed in the main manuscript that the visual prompts in CP-DETR-L are significantly better than those in CP-DETR-T, so further scaling up of the model and training data is still necessary.

\subsection{Visualizations}
In this subsection, we demonstrate the generalisation capabilities of CP-DETR on various scenarios through qualitative visualisations.
In figure~\ref{append2}, we visualise some zero-shot results through textual descriptions.
Our model performs well in different scenarios and correctly processes descriptive text, such as the second row and second column in figure~\ref{append2}.

\begin{figure*}[!ht]
\centering
\includegraphics[scale=0.6]{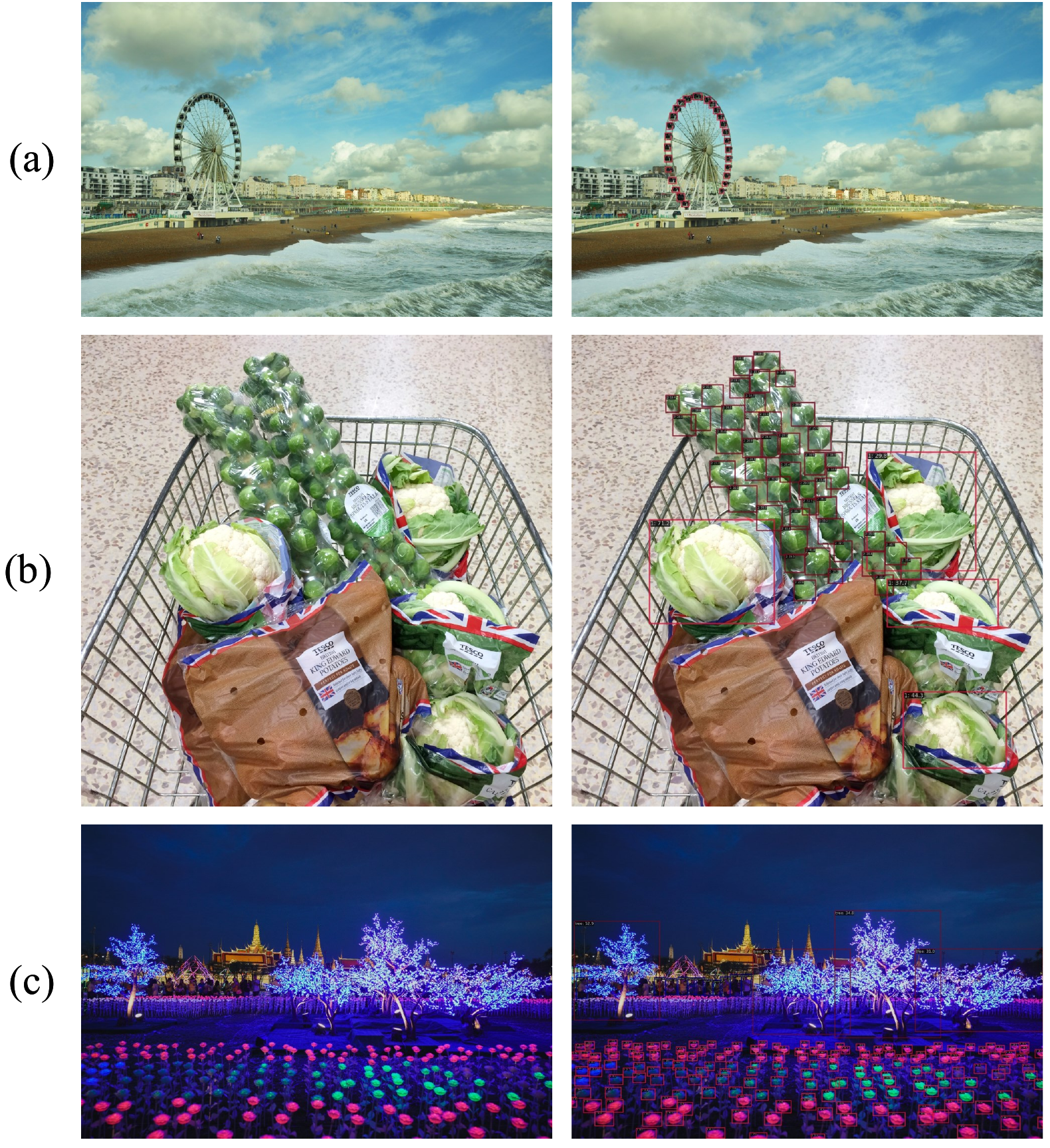} % 
\caption{Visualizations of CP-DETR-L visual prompt outputs. Row 1 use of a class of boxes as inputs.
Row 2 use of two classes of boxes as inputs.
Row 3 use of a class of boxes and text "person.tree" as inputs.}
\label{append3}
\end{figure*}

In figure~\ref{append3}, we visualise some visual prompt results.
It can be observed that visual prompts perform well on dense objects and can be combined with text prompts, as shown in (c) of figure~\ref{append3}.

\subsection{Large-scale model}
Recently, we tried to scale up the model parameters by updating the visual backbone network.
After preliminary experiments, we found that the pre-training weights of the backbone network have a significant effect on the zero-shot performance.
We tried EVA-02~\cite{eva} and Florence-2~\cite{florence} and finally chose EVA-02 ViT-L as the visual backbone of CP-DETR-Pro.
In the preliminary experiments, CP-DETR-Pro uses the same training data as CP-DETR-T and is trained for 16 epochs with batchsize 16.
As shown in table \ref{appendet4}, CP-DETR-Pro exhibits an amazing zero-shot generalization capability, which not only exceeds the best metrics of all open-source algorithms, but is also sufficient to compete with closed-source models trained with tens of times closed-source data.

\subsection{About Code}
The open source code needs to be permitted by China Mobile's Ministry of Science and Innovation, and we are working on applying for it. If there are any changes, we will update the arXiv version to publish the link.

\end{document}